# Carbon Price Forecasting with Quantile Regression and Feature Selection

[1]Tianqi Pang, Kehui Tan, [2]Chenyou Fan
School of Artificial Intelligence, South China Normal University
Guangzhou, China
{[1]pangtianqi.scnu@gmail.com, [2]fanchenyou@scnu.edu.cn}

**Abstract**. Carbon futures has recently emerged as a novel financial asset in the trading markets such as the European Union and China. Monitoring the trend of the carbon price has become critical for both national policy-making as well as industrial manufacturing planning. However, various geopolitical, social, and economic factors can impose substantial influence on the carbon price. Due to its volatility and non-linearity, predicting accurate carbon prices is generally a difficult task. In this study, we propose to improve carbon price forecasting with several novel practices. First, we collect various influencing factors, including commodity prices, export volumes such as oil and natural gas, and prosperity indices. Then we select the most significant factors and disclose their optimal grouping for explainability. Finally, we use the Sparse Quantile Group Lasso and Adaptive Sparse Quantile Group Lasso for robust price predictions. We demonstrate through extensive experimental studies that our proposed methods outperform existing ones. Also, our quantile predictions provide a complete profile of future prices at different levels, which better describes the distributions of the carbon market.

## 1. Introduction

Global scientists and governments have long been concerned about the climate change issue. The global warming effect, as the primary concern of climate change, is mainly caused by the unconstrained emissions of greenhouse gases (GHGs), mainly composed of Carbon Dioxide ($CO_2$), Methane ($CH_4$), and others [1]. Countering global warming has widely become a consensus of human beings, as it profoundly affects the health and sustainability of mankind on Earth.

To tackle this grave issue, many countries and regional unions have established carbon markets to regulate the rights and amounts of carbon emissions. The European Union opened the influential Emission Trading Scheme (EU ETS) in 2005 and has become the worldwide carbon trading place. Since then, carbon trading has become an effective way of regulating the carbon emissions and allocating energy sources [2] in various countries and districts such as China and Australia. Carbon futures has recently emerged as a new financial asset in the trading market. However, the carbon price is influenced by various factors, such as geopolitician, society and economy. Due to its volatility and non-linearity, predicting carbon price accurately is generally difficult. Therefore, developing an effective model that can predict the carbon price has become an emergent research topic. Accurate predictions of carbon price can reduce manufacturing companies' costs through optimal planning and provide insights to governments to regulate the domestic industrial sectors.

In this study, we carry on carbon price forecasting for both EU and Chinese markets. We based our methods mainly on the Quantile Group Lasso (L-QG), which is proposed by Ren and Duan [3]. We optimized and improved the L-QG method. This method selects the optimal factor by applying minimal shrinkage to the quantile array, thereby predicting the dynamic distribution of carbon price returns. However, we observed that many factors are irrelevant such as sparse and power weight, *etc*. Therefore, we propose to improve existed L-QG by adding two penalty terms for ensuring sparsity of factors in groups and assigning adaptive weights to them. We named these two methods as Sparse Quantile Group Lasso (L-SQG) and Adaptive Sparse Quantile Group Lasso (L-ASQG), respectively. For L-SQG, we added sparsity at both per-parameter level with L1-penalty and per-group level with L2-penalty with the parameter $\alpha$ controlling the relative importance of each level. For L-ASQG, we further added adaptivity parameters to improve the flexibility of the model by assigning different penalty weights.

Our contributions can summarize as follows:
(1) We combine the existing Lasso method with the Quantile method for carbon price prediction, and optimized the Quantile Group Lasso [3] to obtain two new methods, L-SQG and L-ASQG.
(2) We develop a multi-factor time-series forecasting method for carbon price prediction by taking into account many related predictors which may influence the carbon market dynamics and automatically select sparse and critical factors out of them.
(3) We perform extensive empirical studies on both carbon markets of the EU and Guangzhou, China. Both datasets have a large number of variables. We study how to select the most significant features and how to group related ones with our proposed Group Lasso variants.

We organize the rest of the paper as follows. Section 2 summarizes the background and existing works of carbon price forecasting and provides the dataset details of EU and Chinese carbon markets. In Section 3, we introduce the Lasso family of regression models, and describe the L-SQG and L-ASQG that we use in our predictions. In Section 4, we carry out extensive experiments to empirically verify our methods on two realistic datasets of EU and Chinese markets. In Section 5 and 6, we provide implementation details and summarize our contributions.

## 2. Related Works

Existing research on carbon price forecasting can mainly divide into two categories: the first is time-series-modeling approaches, and the second is multi-factor-based approaches.

The representative time-series-modeling forecasting methods include GARCH [4], VDM [5], autoregressive moving average (ARIMA) models [6], *etc*. However, they often need to ensure the integrity and smoothness of data and fill in missing values, which are not in most realistic scenarios. Thus, these methods suffer from less reliability. Some methods apply classical machine learning models such as support vector machines (SVMs) [7], Random Forest (RF) [8], XGBoost [9], Multilayer Perceptron (MLP) [10], support vector regression (SVR) [10], and Extreme Learning Machines (ELM) [12] for price prediction. However, these methods need complex feature engineering and largely ignored the potential changes in carbon prices under different price quantiles. In this study, we consider better predicting the distributions of carbon price changes at different quantiles and selecting more significant features.

The other main category of carbon price prediction is based on multi-factors which use hand-designed influencing features and historical data for prediction. Most of these methods utilize ensemble methodology to improve their nonlinearity capacity and performance. Zhu *et al*. [13] developed a general carbon price prediction framework based on decomposition-synthesis which integrates ARIMA, ANN and GAANN models. Zhao *et al*. [14] proposed the decomposition ensemble Empirical Model Decomposition with Adaptive Noise (CEEMDAN) and combined the deep learning method LSTM to project carbon prices. Zhou *et al*. [15] first used the HP filter proposed by Hodrick and Prescott [16] to decompose multi-dimensional data to capture both long-term trends and short-term fluctuations.

## 3. Our Approach

In this part, we briefly introduce the method of regression prediction of carbon futures prices based on Lasso and Quantile Regression. Firstly, we give a brief review of the basic Lasso (original), adaptive Lasso, and group Lasso. We then introduce the Quantile Group Lasso (L-QG) method proposed by Ren and Duan *et al*. [3]. Finally, we propose our Sparse Quantile Group Lasso (L-SQG) and the Adaptive Sparse Quantile Group Lasso (L-ASQG) used in our experiments are introduced.

### 3.1. Least squares regression

Least squares regression, or linear regression, provides an estimate of the conditional mean of the response variable as a function of the covariate. The solution is obtained by minimizing the risk function:

$$R(\beta) = \frac{1}{n}\sum_{i=1}^{n}(y_i - x_i^t\beta)^2. \quad (1)$$

### 3.2. Quantile Regression

Quantile regression initially proposed by Koenker and Bassett [17], focuses on the response of conditional median or conditional quantile information. Quantile regression models how predictor effects vary across response quantile levels. The solution can be found by minimizing the risk function:

$$R(\beta,\tau) = \frac{1}{n}\sum_{i=1}^{n}\rho_\tau(y_i - x_i^t\beta), \quad (2)$$

where $\rho_\tau(u)$ is defined as $\rho_\tau(u) = u(\tau - \mathrm{I}(u \leq 0))$ with $I$ as an indicator function. Intuitively, underestimating the price $y_i > x_i^t\beta$ with high quantile values ($> 0.5$) results in higher penalties than overestimation due to the asymmetrical property of the quantile loss function(2).

### 3.3. Penalized Regression

The carbon price prediction is generally considered to be dependent on many factors such as commodity price and prosperity indices, *etc*. To deal with such high-dimensional data, we describe how we use PCA, Lasso and other methods for dimensionality reduction to prevent overfitting problems caused by excessive data dimensionality. Next, we'll cover some of the existing Lasso methods.

*3.3.1. Sparse Quantile group Lasso (L-SQG)*

Adding sparsity ideas to the previous L-QG, our L-SQG method provides both per-parameter and per-group sparsity and balances terms from Lasso and Group Lasso.

The Sparse Group Lasso estimator is defined as：

$$\hat{\beta} = \arg\min_{\beta \in \mathbb{R}^p}\left\{R(\beta,\tau) + \alpha\lambda\|\beta\|_1 + (1-\alpha)\lambda\sum_{l=1}^{K}\sqrt{p_l}\|\beta^l\|_2\right\}, \quad (3)$$

where $\alpha$ balances sparsity to per-parameter level (L1- penalty) and per-group level (L2-penalty).

We use the L-SQG method to forecast the return on carbon price under the median quantile scenario:

$$\hat{\beta}_{t+1} = \hat{\beta}_{\tau_t,0} + \sum_{i=1}^{N}\hat{\beta}x_{i,t}, \quad (4)$$

Now we can estimate the best regressor by observations of past T months with following equations:

$$\hat{\beta}_{\tau_t,0} = \arg\min_{\beta \in R^p}\left\{\frac{1}{t-1}R(\beta,\tau) + \alpha\lambda\|\beta\|_1 + (1-\alpha)\lambda\sum_{l=1}^{K}\sqrt{p_l}\|\beta^l\|_2\right\}, \quad (5)$$

$$R(\beta) = \frac{1}{n}\sum_{t=1}^{T-1}\left(\rho_{\tau_t}(p_{t+1} - \beta_0 - \sum_{l=1}^{K}\beta_l x_{i,l})\right), \quad (6)$$

where $\hat{\beta}_{\tau_t,0}$ is the regression coefficient estimated by L-SQG using data of $\tau = 0.5$ up to $T$ month, $\beta^l$ is the parameter of the $l$-*th* group, and $\|\beta^l\|_2 = (\sum_{j=1}^{N_l}\beta_l^2)^{\frac{1}{2}}$. $p_{t+1}$ is the logarithmic rate of return of the monthly carbon price, $x_{i,t}$ is the $i$-*th* predictor variable, and $p_i$ is the number of parameters of the $i$-*th* group. When Equation (3) does not have a third term, it is the Lasso (original) method.

*3.3.2. Adaptive sparse Quantile group Lasso (L-ASQG)*

The L-ASQG method further adds adaptative weights on the basis of L-SQG. L-ASQG enhances the flexibility of the model by changing the weights of penalty terms for per-parameter sparsity and per-group sparsity, and improves the accuracy of variable selection and model prediction.

The Adaptive Sparse Group Lasso estimator is defined as:

$$\hat{\beta} = \arg\min_{\beta \in \mathbb{R}^p}\left\{R(\beta,\tau)+\alpha\lambda\sum_{j=1}^{p}\tilde{w}_j|\beta_j|+(1-\alpha)\lambda\sum_{l=1}^{K}\sqrt{p_l}\tilde{v}_l\|\beta^l\|_2\right\}, \quad (7)$$

Where $\tilde{w} \in \mathbb{R}^p$ and $\tilde{v} \in \mathbb{R}^K$ are known weight vectors, and $\tilde{w}_j = 1/|\hat{\beta}_j|^{\gamma 1}$, $\tilde{v}_l = 1/\|\hat{\beta}_l\|_2^{\gamma 2}$, $\gamma 1$ and $\gamma 2$ is a non-negative constant in $[0,2]$. $R(\beta,\tau)$ is the quantile regression defined in Equation (2), $\beta^l$ is the number of parameters in $l$-th group and is used to balance the size difference between groups. In general, when a variable (set of variables) is important, we give $\tilde{v}_l$ ($\tilde{w}_j$) a smaller weight; when a variable (set of variables) is unimportant, we give $\tilde{v}_l$ ($\tilde{w}_j$) a larger weight to penalize it.

We use the L-ASQG method to forecast the return on carbon price under the median scenario:

$$\hat{\beta}_{t+1} = \hat{\beta}_{\tau_t,0} + \sum_{i=1}^{N}\hat{\beta}x_{i,t}, \quad (8)$$

Now we can estimate the best regressor by observations of past $T$ months with following equations:

$$\hat{\beta}_{\tau_t,0} = \arg\min_{\beta \in R^p}\left\{\frac{1}{t-1}R(\beta)+\alpha\lambda\sum_{j=1}^{p}\tilde{w}_j|\beta_j|+(1-\alpha)\lambda\sum_{l=1}^{K}\sqrt{p_l}\tilde{v}_l\|\beta^l\|_2\right\}, \quad (9)$$

$$R(\beta) = \frac{1}{n}\sum_{t=1}^{T-1}\left(\rho_{\tau_t}(p_{t+1}-\beta_0-\sum_{i=1}^{N}\beta_i x_{i,l})\right), \quad (10)$$

where $\hat{\beta}_{\tau_t,0}$ is the regression coefficient estimated by L-ASQG using data of $\tau = 0.5$ up to $T$ month, $\beta^l$ is the parameter of the $l$-th group, the remaining parameters are defined as before. When Equation (7) does not have the second term, it is the Quantile Group Lasso (L-QG) method.

### 3.4. Model evaluation

The model evaluation indicators we use are the Mean Squared Error (MSE), the Mean Absolute Error (MAE), the Root Mean Square Error (RMSE), and the Mean Absolute Percentage Error (MAPE). These are commonly used evaluation metrics for time series forecasting models, and we express them in the Equation (11)-(14). where $y$ is the true value, $\hat{y}$ is the predicted value of the model, $\hat{y}_i$ is the predicted value of the $i$-th sample of the model, and $n$ is the number of samples.

$$MSE(y,\hat{y}) = \frac{1}{n}\sum_{i=1}^{n}(y_i-\hat{y}_i)^2, \quad (11) \qquad MAE(y,\hat{y}) = \frac{1}{n}\sum_{i=1}^{n}|y_i-\hat{y}_i|, \quad (12)$$

$$RMSE(y,\hat{y}) = \sqrt{\frac{1}{n}\sum_{i=1}^{n}(y_i-\hat{y}_i)^2}, \quad (13) \qquad MAPE(y,\hat{y}) = \frac{100\%}{n}\sum_{i=1}^{n}\left|\frac{y_i-\hat{y}_i}{y_i}\right|, \quad (14)$$

For MSE, MAE, and RMSE, their values represent the absolute error between the true and predicted values. For MAPE, it represents the percentage of error, we fill the NULL values with the values of the latter items to ensure that the MAPE equation is available. All metrics are the smaller the better.

## 4. Experiments

### 4.1. Carbon markets and data

The European Union (EU) and China play the leading roles over the world in marketizing the carbon futures. For EU carbon price data, we utilize the carbon futures returns in the EU ETS such as ICE ECX EUA, following [3]. We formulate the target closing price movement as the log of changes of two consecutive monthly closing prices. Figure 1 shows the trend of carbon price trading in the EU from

2009-03 to 2020-12 with a total of 182 data samples. The data splits to training set ranging from 2009-03 to 2019-12 and testing set ranging from 2020-01 to 2020-12. At the same time, the characteristic variables include 44 influencing factors of carbon futures and 18 technical indicators.

For China national carbon price data, we studied Guangdong piloting regional carbon data collected from 2014-2021 for trading Chinese Emission Allowance (CEA). Figure 2 shows the trend of carbon price trading in China from 2009-03 to 2020-12. We split the Guangdong data to training set ranging from 2017-01 to 2020-12, with a total of 1082 samples, and testing set from 2021-01 to 2021-08 with a total of 162 samples. We selected 24 influencing factors as feature vectors, including CSI 300 Index, China Diesel Wholesale Price, *etc*.

### 4.2. Statistics of EU Carbon Prices

We performed the Augmented Dickey–Fuller (ADF) test, Jarque–Bera test, the Autocorrelation Function (ACF), as well as the Partial Autocorrelation Function (PACF). Firstly, we used the ADF test to detect whether the time series data is stable. Generally, a smaller p-value and a more negative test value represent the smoother data. The ADF test value for the EU data is -13.013, the p-value is about 0.00, and the test value at the 5% significance level is -2.882 which is greater than the ADF test value, indicating that the time series is stable. Second, we take advantage of Jarque–Bera Test to determine whether a time series conforms to a normal distribution. The test value is less than 0.05, with skewness and kurtosis as -0.624 and 4.771, respectively. This indicates that the data does not meet the assumption of normality. However, we can still explicitly predict their quantile values with quantile regression. Finally, the ACF and PACF functions are used to determine the autocorrelation of the time series.

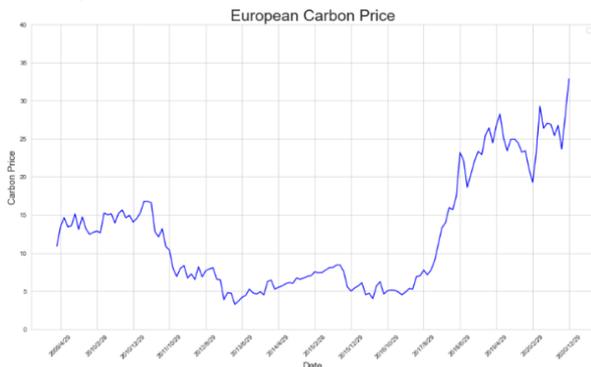 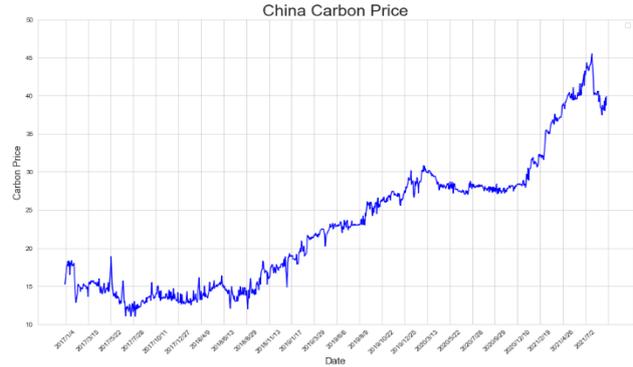

**Figure 1**.Carbon prices of the EU ETS    **Figure 2**.Carbon prices of the Guangzhou ETS

**Table 1.** Statistical tests for the Guangzhou ETS carbon prices

| Categories | Functions | P-value | Hypothesis | implication |
|---|---|---|---|---|
| ADF | adfuller | 0.00 | Reject | Stationarity |
| Jarque_Bera test | Jarque_Bera | 0.00 | Reject | Non-normality |
| ACF | plot_acf | N/A | N/A | Autocorrelation |
| PACF | plot_pacf | N/A | N/A | autocorrelation |

### 4.3. Statistics of Chinese Carbon Prices

As shown in Table 1, the static, normal, and correlation tests are performed for the data in China. First, the ADF test value is -16.054, and the test value at a significance level of less than 5% is -2.864, and the p-value is about 0.00, so the data is stable. Secondly, the Jarque-Bera test yields a p-value of 0.00 less than 0.05 and a skewness of 0.263, kurtosis is 5.474, so the data does not follow a normal distribution. Finally, ACF and PACF plots are obtained. The China data is self-collected from Guangzhou ETS and Chinese economics statistics. We performed a similar feature selection technique to obtain nine critical features according to their importance, including the opening price of

Guangzhou_ETS, the highest price of Guangzhou_ETS, the opening price of the CSI 300, the lowest price, and the ex-factory price of China's LNG, *etc*. As shown in Table 2, We list the feature vectors for all feature selections in EU and Guangzhou.

### 4.4. Feature selection

We now describe the features that are available in the EU and China data. Among them, we select the features with the highest feature importance as predictors which are used in model training.

The EU carbon future data [4] includes factors of crude oil and natural gas production, imports and exports of European countries, futures prices in the market. It also includes plenty of economic indices such as FTSE100 index, CAC 40, M2 values in different countries, unemployment rate, inflation rate, short-/long-term interest rates, *etc*. After using our Lasso-based methods for feature selection, 13 feature vectors are finally selected according to the importance of the features. Important among them are Germany's unemployment rate, the Long-term government bond yield in France, Short-term interest rate in the UK, Unemployment rate in Italy and US natural gas futures, *etc*.

**Table 2.** Feature selection summary

| EU_ETS | Guangzhou_ETS |
|---|---|
| USGF/UKUR | ETS_open(high) |
| GEUR/FRUR | 300 index_close |
| ITIF/GEIF | 300 index_open |
| ITUR/FRIF | 300 index_high |
| UKIR(S/L) | China_diesel_price |
| USIR(L) | 500 index_close |
| FRYD(L) | 500 index_open |
| ITYD(L) | 500 index_high |

**Table 3.** Different methods of parameter Settings

| Paramaters | $\lambda$ | $\alpha$ | Lasso_weight | gl_weight | group_index |
|---|---|---|---|---|---|
| Lasso | √ | -- | -- | -- | -- |
| alasso | √ | -- | √ | -- | -- |
| L-QG | √ | -- | -- | -- | √ |
| L-SQG | √ | √ | -- | -- | √ |
| L-ASQG | √ | √ | √ | √ | √ |

### 4.5. Parameter selection and introduction

As Table 3, L-SQG is the Sparse Quantile Group Lasso method as shown in Equation (3), and L-ASQG is Adaptive Sparse Quantile Group. The Lasso method is shown in Equation (7). The parameter $\lambda$ is regularization parameters used to control the weight of the penalty, the parameter $\alpha$ regulates the balance between Lasso and group Lasso. The parameters Lasso_weight ($\tilde{w}_j$) correspond to the method of adaptive weight adjustment. The parameters gl_weight ($\tilde{v}_l$) need to be set to adjust the weight between groups. The parameter group_index corresponds to gl_weight, and features are grouped at the same time as gl_weight uesd. We discuss these parameters in details in Section 5.

### 4.6. Results analysis

We compare the newly improved methods L-SQG and L-ASQG with several other Lasso-based methods on the EU dataset and the Guangzhou dataset in China, and the Tables 4 and 5 show the results. Firstly, our proposed method L-ASQG has the lowest MSE (0.0194), MAE (0.1241), RMSE (0.1393) and MAPE (2.6600) compared to other Lasso-based models on the European ETS carbon price dataset. Therefore, compared with the L-QG [3], the results obtained by us are improved by about 0.01%~0.04% under the same dataset and parameters. This indicates that the Quantile Group Lasso model will perform better after adding adaptive and sparsity terms simultaneously. Secondly, the results of the L-SQG method proposed by us are basically not better than L-QG, indicating that only adjusting the sparsity for per-parameter and per-group levels cannot improve the results obviously unless the data has larger dimensions.

Then, on the carbon price dataset of ETS in Guangzhou, China, similar to the EU dataset, our proposed L-ASQG model also has the lowest MSE (0.0362), MAE (0.1495), RMSE (0.1903) and MAPE (3.0894). We found that the improvement effect was not as good as that of the EU ETS dataset because the number of feature vectors was too small. Therefore, when the feature selection is carried out by the Lasso method, the proportion of features with high importance is relatively small, and the correlation between the groups is not as high as that of the EU ETS dataset when grouping the features, which will lead to the reduction of the influence of parameter $\alpha$, parameter Lasso_weight and parameter gl_weight.

For both the lm_Lasso model and the qr_Lasso model, it is clear that qr_Lasso performs significantly better than lm_Lasso on both datasets, indicating that for high-dimensional time series models, quantile regression is significantly better than using only linear regression (least squares). Especially for nonlinear time series data such as carbon price. This proves that it is suitable to use quantile regression to solve the problem. Finally, the alasso model is also an improvement over the qr_Lasso model, which also includes the idea of quantile regression, combined with the idea of adaptation, gives a weight value to the penalty term, so that the prediction effect is further improved.

Table 4. EU Carbon Price Different Model Evaluation Indicators

| Assessment indicators | MSE | MAE | RMSE | MAPE |
| --- | --- | --- | --- | --- |
| lm_Lasso | 0.0379 | 0.1724 | 0.1948 | 4.4492 |
| qr_Lasso | 0.0259 | 0.1359 | 0.1608 | 2.7713 |
| aLasso | 0.0254 | 0.1348 | 0.1593 | 2.7128 |
| L-QG | 0.0195 | 0.1245 | 0.1396 | 2.6845 |
| L-SQG | 0.0195 | 0.1245 | 0.1396 | 2.6845 |
| L-ASQG | **0.0194** | **0.1241** | **0.1393** | **2.6600** |

Table 5. Guangzhou Carbon Price Different Model Evaluation Indicators

| Assessment indicators | MSE | MAE | RMSE | MAPE |
| --- | --- | --- | --- | --- |
| lm_Lasso | 0.0507 | 0.1777 | 0.2252 | 4.4325 |
| qr_Lasso | 0.0382 | 0.1543 | 0.1955 | 3.3311 |
| alasso | 0.0136 | 0.0912 | 0.1168 | 4.2450 |
| L-QG | 0.0362 | 0.1496 | 0.1904 | 3.0920 |
| L-SQG | 0.0362 | 0.1496 | 0.1904 | 3.0920 |
| L-ASQG | **0.0362** | **0.1495** | **0.1903** | **3.0894** |

### 4.7. Discussions of results under different quantile

Since there are too few forecasts for EU, we only compare the distribution of true and predicted values for China at different quantiles. Tables 6 represent the results of the evaluation indicators of the three methods of L-QG, L-SQG and L-ASQG at quantiles of 0.25, 0.5 and 0.75, respectively. Figure 3 represent the time series plots of the method of L-ASQG at quantiles of 0.25, 0.5 and 0.75, respectively. Among them, the blue line represents the true value of Guangzhou ETS Carbon price from 2021-1-3 to 2021-8-31, the red line represents the predicted value when the quantile $\tau = 0.25$, the green line represents the predicted value when the quantile is 0.5, and the yellow line represents the predicted value when the quantile is 0.75.

Table 6. Guangzhou Carbon Price Different Model Evaluation Indicators ($\tau$=0.25/0.5/0.75)

| Method | L-QG | L-SQG | L-ASQG |
| --- | --- | --- | --- |
| MSE | 0.0760/0.0427/0.0362 | 0.0760/0.0427/0.0362 | **0.0759**/0.0427/0.0362 |
| MAE | 0.2155/0.1582/0.1496 | 0.2155/0.1582/0.1496 | **0.2153**/0.1582/**0.1495** |
| RMSE | 0.2757/0.2067/0.1904 | 0.2757/0.2067/0.1904 | **0.2754**/0.2067/**0.1903** |
| MAPE | 5.9330/4.6412/3.0920 | 5.9330/4.6412/3.0920 | **5.9240/4.6392/3.0894** |

First, we observe that the model has the best prediction performance when $\tau = 0.5$, and the values of all evaluation indicators are the lowest. The effect of each evaluation index at $\tau = 0.25$ was better than that at $\tau = 0.75$. Secondly, when $\tau = 0.25$, the L-ASQG model only has a slight improvement in the results of MAPE. When $\tau = 0.5$, except for the value of MSE, the results of other evaluation indicators were improved. When $\tau = 0.75$, we find that all indicators have improved more than before. This shows that as the quantile increases, the prediction effect of L-ASQG gradually increases. Quantile regression differs from linear regression as the quantile regression minimizes the sum of absolute errors produced by the selected quantile tangent points instead of squared errors. According to the different values of $\tau$ we define, the variation trend of the obtained explanatory variable $x$ to the quantile of the response variable $y$ is different, and the degree of influence is also different.

## 5. Implementation Details

We utilize Python and Sklearn to implement plain Lasso. We also use the python package "asgl" [18] to implement alasso, gl, sgl, asgl and make fair comparisons. First, we perform logarithmic differential processing on the dataset to represent the carbon price yield, which can make the dataset more stable on the one hand, and improve the accuracy of prediction on the other hand. After logarithmic differentiation, there are many NaN values and INF values, and we use the mean of the test set to process the NaN and INF values to ensure that the error is further reduced. Second, we are based on Python3.8 8 and the Lasso function in the sklearn library to feature the variables. A total of 9 feature vectors with high importance were selected in the data of China, and the parameter group_index = [1,1,1,1,1,1,1,2,2]. A total of 13 feature vectors were selected for the EU data, and the parameter group_index = [1,1,1,1,1,1,1,2,2,2,2,2,2,2,2,2]. We use the ASGL library to fit the parameter Lasso_weight ($\tilde{w}_j$) and parameter gl_weight ($\tilde{v}_l$). Then we use the CV (cross-validation) function in the ASGL library to search for the optimal parameter values. We search a parameter space of $\lambda$ of 10 to the power of t with t in [-5,1.01] with step size of 0.2, including 31 numeric values. We set $\alpha$ =0.0001, power_weight = [-0.4, -0.2, 0, 0.2, 0.4]. The power_weight is used to compute the Lasso_weight and gl_weight.

### 5.1. Discussions about data differences

The main differences between Chinese data and EU data are as follows:
(1) Data for the EU region are analyzed on a monthly basis and are on the small side, but there are 44 influencing factors and 18 technical indicators.
(2) The data for China is analyzed on a daily basis, which is larger than in EU, but it is also a small sample data set with a total of 24 contributing factors.
(3) The carbon price in EU is futures, but the carbon price in China is spot goods. This is because China's carbon trading institutions were established late, and there is currently only a spot carbon market.

## 6. Conclusion

In this paper, two Lasso-based quantile regression methods, L-SQG and L-ASQG, are summarized, and their validity and stability are demonstrated by verification and comparison on European datasets and Chinese datasets. The main conclusions are as follows:
(1) The L-ASQG model combining sparsity and adaptivity thinking can improve the accuracy of carbon price forecasting. And it performs best under different quantile conditions.

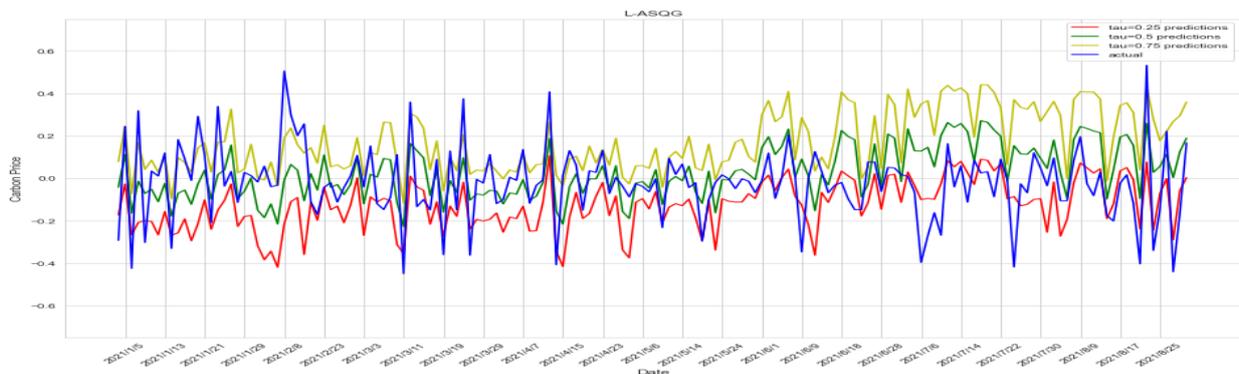

**Figure 3**. L-ASQG Model Guangzhou ETS carbon price distribution under different quantiles

(2) Our methods are more suitable for small-sample high-dimensional data models and conforms to the data characteristics of the carbon market.

**References**

[1] Adedoyin, F.F., Gumede, M.I., Bekun, F.V., Etokakpan, M.U., Balsalobre-lorente, D., 2020. Modelling coal rent, economic growth and CO2 emissions: does regulatory quality matter in BRICS economies? Sci. Total Environ. 710, 136284.
[2] Dutta, A., 2016. Modeling and forecasting the volatility of carbon emission market: the role of outliers, time-varying jumps and oil price risk. J. Clean. Prod. 172, 2773–2781.
[3] Xiaohang Ren, Kun Duan, Lizhu Tao, Yukun Shi, Cheng Yan, Carbon prices forecasting in quantiles. 2022. Energy Economics, Vol. 108.
[4] Engle, R., 2001.GARCH 101: The use of ARCH/GARCH models in applied econometrics. J. Econ. Perspect. 15 (4), 157–168.
[5] Sun, G., Chen, T., Wei, Z., Sun, Y., Zang, H., Chen, S., 2016. A carbon price forecasting model based on variational mode decomposition and spiking neural networks. Energies 9 (1), 54.
[6] Mohammadi H, Su L (2010) International evidence on crude oil price dynamics: applications of ARIMA-GARCH models. Energy Econ 32:1001–1008.
[7] Zhu B, Han D, Wang P et al (2017) Forecasting carbon price using empirical mode decomposition and evolutionary least squares support vector regression. Appl Energy 191:521–530.
[8] Mei J, et al (2014) A random forest method for realtime price forecasting in New York electricity market. In: 2014 IEEE PES General Meeting Conference & Exposition. IEEE, pp1–5.
[9] Gumus M, Kiran MS (2017) Crude oil price forecasting using XGBoost. In: 2017 International conference on computer science and engineering (UBMK). IEEE, pp 1100–1103.
[10] Tatar, A., Naseri, S., Bahadori, M., Hezave, A.Z., Kashiwao, T., Bahadori, A., Darvish, H., 2016. Prediction of carbon dioxide solubility in ionic liquids using MLP and radial basis function (RBF) neural networks. J. Taiwan Inst. Chem. Eng. 60, 151–164.
[11] Fan, X., Li, S., Tian, L., 2015. Chaotic characteristic identification for carbon price and a multi-layer perceptron network prediction model. Expert Syst. Appl. 42, 3945–3952.
[12] Xu, H., Wang, M., Jiang, S., Yang, W., 2020. Carbon price forecasting with complex network and extreme learning machine. Phys. A 545, 122830.
[13] Zhu B. A novel multiscale ensemble carbon price prediction model integrating empirical mode decomposition, genetic algorithm and artificial neural network. Energies 2012;5(2):355–70.
[14] Zhao, L., Miao, J., Qu, S., Chen, X., 2021. A multi-factor integrated model for carbon price forecasting: Market interaction promoting carbon emission reduction. Science of The Total Environment.796(2021), Article 149110.
[15] Zhou, F., Huang, Z., Zhang, C., 2022. Carbon price forecasting based on CEEMDAN and LSTM. Applied Energy.311(2022), Article 118601.
[16] Prescott, R.J..H.E.C., 2016. Postwar U.S. Business Cycles  An Empirical Stablel. 29. Ohio State Univ. Press, pp. 1–16.
[17] R. Koenker and G. Bassett, 2005. Regression Quantiles. Econometrica, 46(1):33–50, 1 1978. ISSN 00129682.
[18] Mendez-Civieta, M. C. Aguilera-Morillo, and R. E. Lillo. Adaptive sparse group Lasso in quantile regression. Advances in Data Analysis and Classification, 2020. ISSN 18625355.